\documentclass{article}
\usepackage{spconf,amsmath,bm,graphicx}
\usepackage[utf8]{inputenc}
\usepackage{pgf, tikz}
\usetikzlibrary{arrows, fit, calc, shapes.geometric,positioning, automata, petri, topaths}

\DeclareGraphicsExtensions{.pdf,.jpeg,.png}

\title{LONG SHORT-TERM MEMORY BASED RECURRENT NEURAL NETWORK ARCHITECTURES FOR LARGE VOCABULARY SPEECH RECOGNITION}

\name{Haşim Sak, Andrew Senior, Françoise Beaufays}
\address{Google\\ {\small \tt\{hasim,andrewsenior,fsb@google.com\}}}

\begin{document}
\ninept

\maketitle
\begin{abstract}
Long Short-Term Memory (LSTM) is a recurrent neural network (RNN) architecture that has been designed to address the vanishing and exploding gradient problems of conventional RNNs. Unlike feedforward neural networks, RNNs have cyclic connections making them powerful for modeling sequences.
They have been successfully used for sequence labeling and sequence prediction tasks, such as handwriting recognition, language modeling, phonetic labeling of acoustic frames.
However, in contrast to the deep neural networks, the use of RNNs in speech recognition has been limited to phone recognition in small scale tasks.
In this paper, we present novel LSTM based RNN architectures which make more effective use of model parameters to train acoustic models for large vocabulary speech recognition.
We train and compare LSTM, RNN and DNN models at various numbers of parameters and configurations.
We show that LSTM models converge quickly and give state of the art speech recognition performance for relatively small sized models.
\footnote{The original manuscript has been submitted to ICASSP 2014 conference on November 4, 2013 and it has been rejected due to having content on the reference only 5th page. This version has been slightly edited to reflect the latest experimental results.}
\end{abstract}


\section{Introduction}
\label{sec:intro}
Unlike feedforward neural networks (FFNN) such as deep neural networks (DNNs), the architecture of recurrent neural networks (RNNs) have cycles feeding the activations from previous time steps as input to the network to make a decision for the current input.
The activations from the previous time step are stored in the internal state of the network and they provide indefinite temporal contextual information in contrast to the fixed contextual windows used as inputs in FFNNs.
Therefore, RNNs use a dynamically changing contextual window of all sequence history rather than a static fixed size window over the sequence.
This capability makes RNNs better suited for sequence modeling tasks such as sequence prediction and sequence labeling tasks.

However, training conventional RNNs with the gradient-based back-propagation through time (BPTT) technique is difficult due to the vanishing gradient and exploding gradient problems~\cite{Bengio:94}.
In addition, these problems limit the capability of RNNs to model the long range context dependencies to 5-10 discrete time steps between relevant input signals and output.

To address these problems, an elegant RNN architecture -- \textit{Long Short-Term Memory} (LSTM) -- has been designed~\cite{Hochreiter:97}.
The original architecture of LSTMs contained special units called \textit{memory blocks} in the recurrent hidden layer.
The memory blocks contain memory cells with self-connections storing (remembering) the temporal state of the network in addition to special multiplicative units called gates to control the flow of information.
Each memory block contains an \textit{input gate} which controls the flow of input activations into the memory cell and an \textit{output gate} which controls the output flow of cell activations into the rest of the network.
Later, to address a weakness of LSTM models preventing them from processing continuous input streams that are not segmented into subsequences -- which would allow resetting the cell states at the begining of subsequences -- a \textit{forget gate} was added to the memory block~\cite{Gers:00}.
A forget gate scales the internal state of the cell before adding it as input to the cell through self recurrent connection of the cell, therefore adaptively forgetting or resetting cell's memory.
Besides, the modern LSTM architecture contains \textit{peephole connections} from its internal cells to the gates in the same cell to learn precise timing of the outputs~\cite{Gers:03}.

LSTMs and conventional RNNs have been successfully applied to sequence prediction and sequence labeling tasks.
LSTM models have been shown to perform better than RNNs on learning context-free and context-sensitive languages~\cite{Gers:01}.
Bidirectional LSTM networks similar to bidirectional RNNs~\cite{Schuster:97} operating on the input sequence in both direction to make a decision for the current input has been proposed for phonetic labeling of acoustic frames on the TIMIT speech database~\cite{Graves:05}.
For online and offline handwriting recognition, bidirectional LSTM networks with a connectionist temporal classification (CTC) output layer using a forward backward type of algorithm which allows the network to be trained on unsegmented sequence data, have been shown to outperform a state of the art HMM-based system~\cite{Graves:09}.
Recently, following the success of DNNs for acoustic modeling~\cite{Mohamed:12, Dahl:12, Jaitly:12}, a deep LSTM RNN -- a stack of multiple LSTM layers -- combined with a CTC output layer and an RNN transducer predicting phone sequences -- has been shown to get the state of the art results in phone recognition on the TIMIT database~\cite{Graves:13}.
In language modeling, a conventional RNN has obtained very significant reduction of perplexity over standard $n$-gram models~\cite{Mikolov:10}.

While DNNs have shown state of the art performance in both phone recognition and large vocabulary speech recognition~\cite{Mohamed:12, Dahl:12, Jaitly:12}, the application of LSTM networks has been limited to phone recognition on the TIMIT database, and it has required using additional techniques and models such as CTC and RNN transducer to obtain better results than DNNs.

In this paper, we show that LSTM based RNN architectures can obtain state of the art performance in a large vocabulary speech recognition system with thousands of context dependent (CD) states.
The proposed architectures modify the standard architecture of the LSTM networks to make better use of the model parameters while addressing the computational efficiency problems of large networks.

\section{LSTM Architectures}
\label{sec:lstm}
In the standard architecture of LSTM networks, there are an input layer, a recurrent LSTM layer and an output layer.
The input layer is connected to the LSTM layer.
The recurrent connections in the LSTM layer are directly from the cell output units to the cell input units, input gates, output gates and forget gates.
The cell output units are connected to the output layer of the network.
The total number of parameters $W$ in a standard LSTM network with one cell in each memory block, ignoring the biases, can be calculated as follows:
$$W = n_c \times n_c \times 4 + n_i \times n_c \times 4 + n_c \times n_o + n_c \times 3$$
where $n_c$ is the number of memory cells (and number of memory blocks in this case), $n_i$ is the number of input units, and $n_o$ is the number of output units.
The computational complexity of learning LSTM models per weight and time step with the stochastic gradient descent (SGD) optimization technique is $O(1)$.
Therefore, the learning computational complexity per time step is $O(W)$.
The learning time for a network with a relatively small number of inputs is dominated by the $n_c \times (n_c + n_o)$ factor.
For the tasks requiring a large number of output units and a large number of memory cells to store temporal contextual information, learning LSTM models become computationally expensive.

\begin{figure}[tb]

\tikzstyle{block} = [draw, rectangle, minimum height=4em, minimum width=1.0em]
\tikzstyle{optblock} = [draw, dashed, rectangle, minimum height=4em, minimum width=1.0em]
\tikzstyle{input} = [coordinate]
\tikzstyle{output} = [coordinate]

\tikzstyle{unit}=[circle,thick,draw,minimum size=4mm]
\tikzstyle{cunit}=[circle,thick,draw,scale=0.6]
\tikzstyle{munit}=[circle,thick,draw,scale=0.5]

\begin{tikzpicture}[auto, ->, node distance=1.32cm, >=latex']
  \node [block, name=input, xshift=2mm] {\rotatebox{90}{input}};

  \node [unit, name=mbinput, right of=input] {$g$};
  \node [munit, name=cellinput, right of=mbinput] {$\bm{\times}$};
  \node [cunit, name=cell, right of=cellinput] {$c_{t-1}$};
  \node [unit, name=celloutput, right of=cell] {$h$};
  \node [unit, name=inputgate, above of=cellinput] {$$};
  \node [munit, name=forgetunit, above of=cell] {$\bm{\times}$};
  \node [unit, name=forgetgate, above=0.5cm of forgetunit] {$$};
  \node [munit, name=mboutput, right of=celloutput] {$\bm{\times}$};
  \node [unit, name=outputgate, above of=mboutput] {$$};
  \draw (mbinput) to node {} (cellinput);
  \draw (inputgate) to node [left] {$i_t$} (cellinput);
  \draw (cellinput) to node {} (cell);
  \draw (forgetgate) to node [left] {$f_t$} (forgetunit);
  \draw [bend right=60] (forgetunit) to node [auto] {} (cell.west);
  \draw (cell.center) to node {} (forgetunit);
  \draw [dotted] (cell.center) to node {} (inputgate.east);
  \draw [dotted, bend right=50] (cell.center) to node {} (forgetgate.east);
  \draw (cell) to node {$c_t$} (celloutput.west);
  \draw [dotted] (cell.center) to node {} (outputgate.west);
  \draw (celloutput) to node {} (mboutput);
  \draw (outputgate) to node [left] {$o_t$} (mboutput);

  \node [block, name=rprojection, xshift=0.5cm, yshift=0.1cm, above right of=mboutput] {\rotatebox{90}{recurrent}};
  \node [optblock, name=projection, xshift=0.5cm, yshift=-0.1cm, below right of=mboutput] {\rotatebox{90}{projection}};
  \node [block, name=output, below right of=rprojection,xshift=2mm] {\rotatebox{90}{output}};
  \node [coordinate, name=result, xshift=-0.25cm, right of=output] {};
  \draw (input) -- node [below] {$x_t$} (mbinput);
  \draw (input) -- ++(0.6cm,0) -- ++(0, 2.49cm) -| (inputgate.north);
  \draw (input) -- ++(0.6cm,0) -- ++(0, 2.49cm) -| (outputgate.north);
  \draw (input) -- ++(0.6cm,0) -- ++(0, 2.49cm) -| (forgetgate.north);
  \draw (mboutput) -- ++(0.8cm,0) node [above left] {$m_t$} |- (rprojection.west);
  \draw [dashed] (mboutput) -- ++(0.8cm,0) |- (projection.west);
  \draw [dashed] (projection.east) -- ++(0.45cm,0) node [above left, xshift=0.5mm] {$p_t$} |- (output.west);
  \draw (rprojection.east) -- ++(0.45cm,0) node [above left, xshift=0.5mm] {$r_t$} |- (output.west);
  \draw (rprojection.east) -- ++(0.45cm,0) -- ++(0, 1.4cm) -| node [near start, above] {$r_{t-1}$} ($(mbinput.north) + (-0.0cm, 0em)$);
  \draw (rprojection.east) -- ++(0.45cm,0) -- ++(0, 1.4cm) -| (inputgate.north);
  \draw (rprojection.east) -- ++(0.45cm,0) -- ++(0, 1.4cm) -| (outputgate.north);
  \draw (rprojection.east) -- ++(0.45cm,0) -- ++(0, 1.4cm) -| (forgetgate.north);
  \draw (output.east) -- node {$y_t$} (result);

  \tikzset{blue dotted/.style={draw=blue!50!white, line width=1pt, dash pattern=on 1pt off 4pt on 6pt off 4pt, inner sep=0.4mm, rectangle, rounded corners}};
  \node (memory blocks) [blue dotted, fit = (mbinput) (mboutput) (cell) (celloutput) (cellinput) (forgetgate) (inputgate) (outputgate)] {};
  \node at (memory blocks.north) [above, inner sep=1mm] {memory blocks};
\end{tikzpicture}

\caption{LSTM based RNN architectures with a recurrent projection layer and an optional non-recurrent projection layer. A single memory block is shown for clarity.}
\label{fig:lstm}
\end{figure}
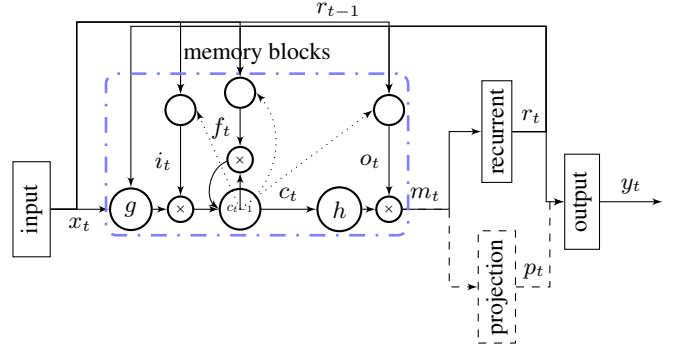

As an alternative to the standard architecture, we propose two novel architectures to address the computational complexity of learning LSTM models.
The two architectures are shown in the same Figure~\ref{fig:lstm}.
In one of them, we connect the cell output units to a recurrent projection layer which connects to the cell input units and gates for recurrency in addition to network output units for the prediction of the outputs.
Hence, the number of parameters in this model is $n_c \times n_r \times 4 + n_i \times n_c \times 4 + n_r \times n_o + n_c \times n_r + n_c \times 3$, where $n_r$ is the number of units in the recurrent projection layer.
In the other one, in addition to the recurrent projection layer, we add another non-recurrent projection layer which is directly connected to the output layer.
This model has $n_c \times n_r \times 4 + n_i \times n_c \times 4 + (n_r + n_p) \times n_o + n_c \times (n_r + n_p) + n_c \times 3$ parameters, where $n_p$ is the number of units in the non-recurrent projection layer and it allows us to increase the number of units in the projection layers without increasing the number of parameters in the recurrent connections ($n_c \times n_r \times 4$).
Note that having two projection layers with regard to output units is effectively equivalent to having a single projection layer with $n_r + n_p$ units.

An LSTM network computes a mapping from an input sequence $x = (x_1,...,x_T)$ to an output sequence $y = (y_1,...,y_T)$ by calculating the network unit activations using the following equations iteratively from $t=1$ to $T$:

\begin{eqnarray}
i_t = \sigma (W_{ix} x_t + W_{im} m_{t-1} + W_{ic} c_{t-1} + b_i) \\
f_t = \sigma (W_{fx} x_t + W_{mf} m_{t-1} + W_{cf} c_{t-1} + b_f) \\
c_t = f_t \odot c_{t-1} + i_t \odot g(W_{cx} x_t + W_{cm} m_{t-1} + b_c) \\
o_t = \sigma (W_{ox} x_t + W_{om} m_{t-1} + W_{oc} c_{t} + b_o) \\
m_t = o_t \odot h(c_t) \\
y_t = W_{ym} m_t + b_y
\end{eqnarray}
where the $W$ terms denote weight matrices (e.g. $W_{ix}$ is the matrix of weights from the input gate to the input), the $b$ terms denote bias vectors ($b_i$ is the input gate bias vector), $\sigma$ is the logistic sigmoid function, and $i$, $f$, $o$ and $c$ are respectively the input gate, forget gate, output gate and cell activation vectors, all of which are the same size as the cell output activation vector $m$, $\odot$ is the element-wise product of the vectors and $g$ and $h$ are the cell input and cell output activation functions, generally $tanh$.

With the proposed LSTM architecture with both recurrent and non-recurrent projection layer, the equations are as follows:
\begin{eqnarray}
i_t = \sigma (W_{ix} x_t + W_{ir} r_{t-1} + W_{ic} c_{t-1} + b_i) \\
f_t = \sigma (W_{fx} x_t + W_{rf} r_{t-1} + W_{cf} c_{t-1} + b_f) \\
c_t = f_t \odot c_{t-1} + i_t \odot g(W_{cx} x_t + W_{cr} r_{t-1} + b_c) \\
o_t = \sigma (W_{ox} x_t + W_{or} r_{t-1} + W_{oc} c_{t} + b_o) \\
m_t = o_t \odot h(c_t) \\
r_t = W_{rm} m_t \\
p_t = W_{pm} m_t \\
y_t = W_{yr} r_t + W_{yp} p_t + b_y \\
\end{eqnarray}
where the $r$ and $p$ denote the recurrent and optional non-recurrent unit activations.

\subsection{Implementation}
\label{sec:impl}
We choose to implement the proposed LSTM architectures on multi-core CPU on a single machine rather than on GPU.
The decision was based on CPU's relatively simpler implementation complexity and ease of debugging.
CPU implementation also allows easier distributed implementation on a large cluster of machines if the learning time of large networks becomes a major bottleneck on a single machine~\cite{Dean:12}.
For matrix operations, we use the Eigen matrix library~\cite{Eigen:10}.
This templated C++ library provides efficient implementations for matrix operations on CPU using vectorized instructions (SIMD -- single instruction multiple data).
We implemented activation functions and gradient calculations on matrices using SIMD instructions to benefit from parallelization.

We use the asynchronous stochastic gradient descent (ASGD) optimization technique.
The update of the parameters with the gradients is done asynchronously from multiple threads on a multi-core machine.
Each thread operates on a batch of sequences in parallel for computational efficiency -- for instance, we can do matrix-matrix multiplications rather than vector-matrix multiplications -- and for more stochasticity since model parameters can be updated from multiple input sequence at the same time.
In addition to batching of sequences in a single thread, training with multiple threads effectively results in much larger batch of sequences (number of threads times batch size) to be processed in parallel.

We use the truncated backpropagation through time (BPTT) learning algorithm to update the model parameters~\cite{Williams:90}.
We use a fixed time step $T_{bptt}$ (e.g. 20) to forward-propagate the activations and backward-propagate the gradients.
In the learning process, we split an input sequence into a vector of subsequences of size $T_{bptt}$.
The subsequences of an utterance are processed in their original order.
First, we calculate and forward-propagate the activations iteratively using the network input and the activations from the previous time step for $T_{bptt}$ time steps starting from the first frame and calculate the network errors using network cost function at each time step.
Then, we calculate and back-propagate the gradients from a cross-entropy criterion, using the errors at each time step and the gradients from the next time step starting from the time $T_{bptt}$.
Finally, the gradients for the network parameters (weights) are accumulated for $T_{bptt}$ time steps and the weights are updated.
The state of memory cells after processing each subsequence is saved for the next subsequence.
Note that when processing multiple subsequences from different input sequences, some subsequences can be shorter than $T_{bptt}$ since we could reach the end of those sequences.
In the next batch of subsequences, we replace them with subsequences from a new input sequence, and reset the state of the cells for them.

\section{Experiments}
\label{sec:exp}
We evaluate and compare the performance of DNN, RNN and LSTM neural network architectures on a large vocabulary speech recognition task -- Google English Voice Search task.

\subsection{Systems \& Evaluation}
\label{sec:eval}
All the networks are trained on a 3 million utterance (about 1900 hours) dataset consisting of anonymized and hand-transcribed Google voice search and dictation traffic.
The dataset is represented with 25ms frames of 40-dimensional log-filterbank energy features computed every 10ms.
The utterances are aligned with a 90 million parameter FFNN with 14247 CD states.
We train networks for three different output states inventories: 126, 2000 and 8000. These are obtained by mapping 14247 states down to these smaller state inventories through equivalence classes.
The 126 state set  are the context independent (CI) states (3 x 42).
The weights in all the networks before training are randomly initialized.
We try to set the learning rate specific to a network architecture and its configuration to the largest value that results in a stable convergence.
The learning rates are exponentially decayed during training.

During training, we evaluate frame accuracies (i.e. phone state labeling accuracy of acoustic frames) on a held out development set of 200,000 frames.
The trained models are evaluated in a speech recognition system on a test set of 23,000 hand-transcribed utterances and the word error rates (WERs) are reported.
The vocabulary size of the language model used in the decoding is 2.6 million.

The DNNs are trained with SGD with a minibatch size of 200 frames on a Graphics Processing Unit (GPU).
Each network is fully connected with logistic sigmoid hidden layers and with a softmax output layer representing phone HMM states.
For consistency with the LSTM architectures, some of the networks have a low-rank projection layer~\cite{Sainath:13}.
The DNNs inputs consist of stacked frames from an asymmetrical window, with 5 frames on the right and either 10 or 15 frames on the left (denoted 10w5 and 15w5 respectively)

The LSTM and conventional RNN architectures of various configurations are trained with ASGD with 24 threads, each asynchronously processing one partition of data, with each thread computing a gradient step on 4 or 8 subsequences from different utterances.
A time step of 20 ($T_{bptt}$) is used to forward-propagate and  the activations and backward-propagate the gradients using the truncated BPTT learning algorithm.
The units in the hidden layer of RNNs use the logistic sigmoid activation function.
The RNNs with the recurrent projection layer architecture use linear activation units in the projection layer.
The LSTMs use hyperbolic tangent activation (tanh) for the cell input units and cell output units, and logistic sigmoid for the input, output and forget gate units.
The recurrent projection and optional non-recurrent projection layers in the LSTMs use linear activation units.
The input to the LSTMs and RNNs is 25ms frame of 40-dimensional log-filterbank energy features (no window of frames).
Since the information from the future frames helps making better decisions for the current frame, consistent with the DNNs, we delay the output state label by 5 frames.

\subsection{Results}
\label{sec:results}

\begin{figure}[ht]
  \centering
  \centerline{\includegraphics[width=8.5cm]{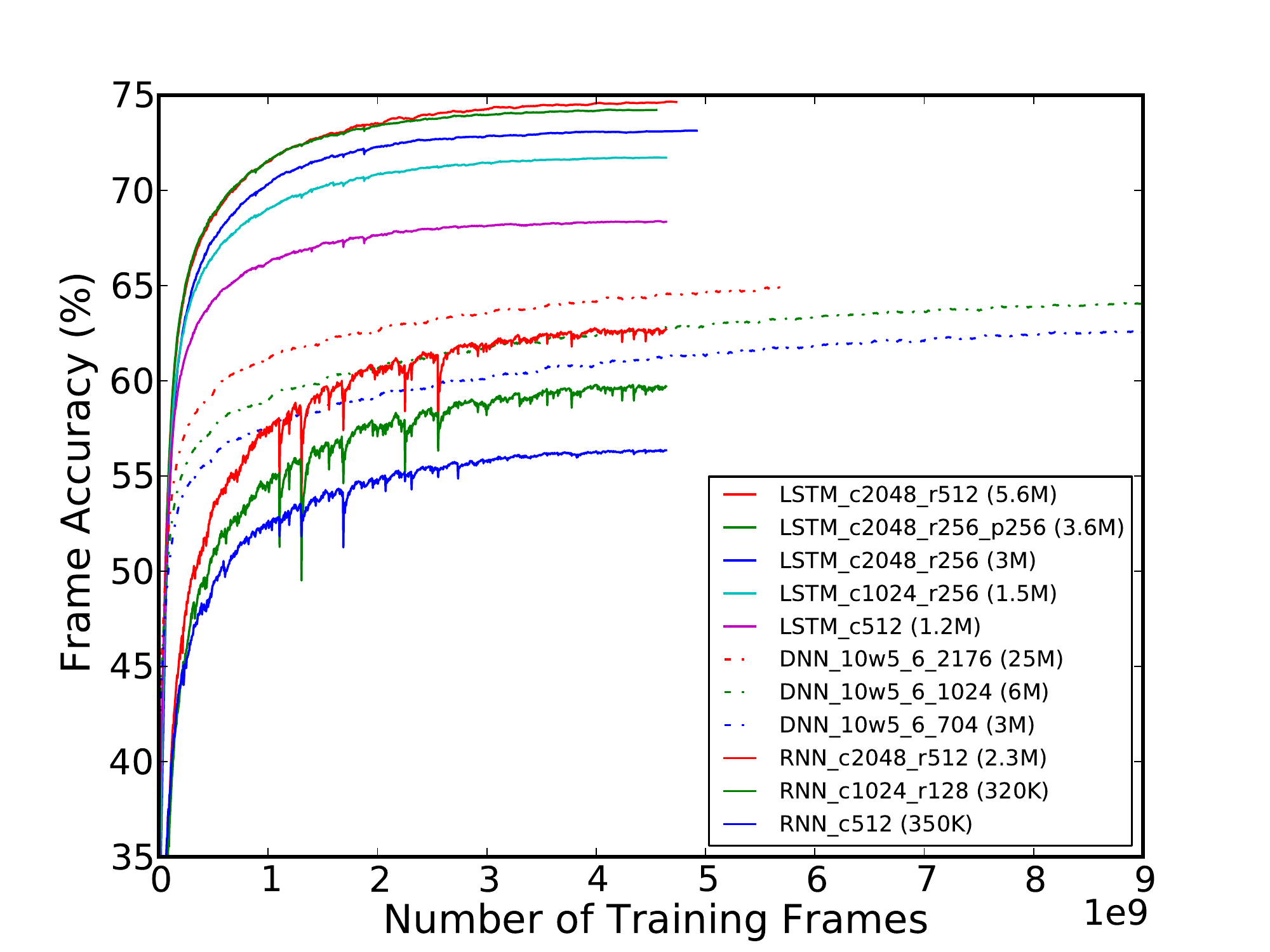}}
\caption{126 context independent phone HMM states.}
\label{fig:frame_acc1}
\end{figure}

\begin{figure}[ht]
  \centering
  \centerline{\includegraphics[width=8.5cm]{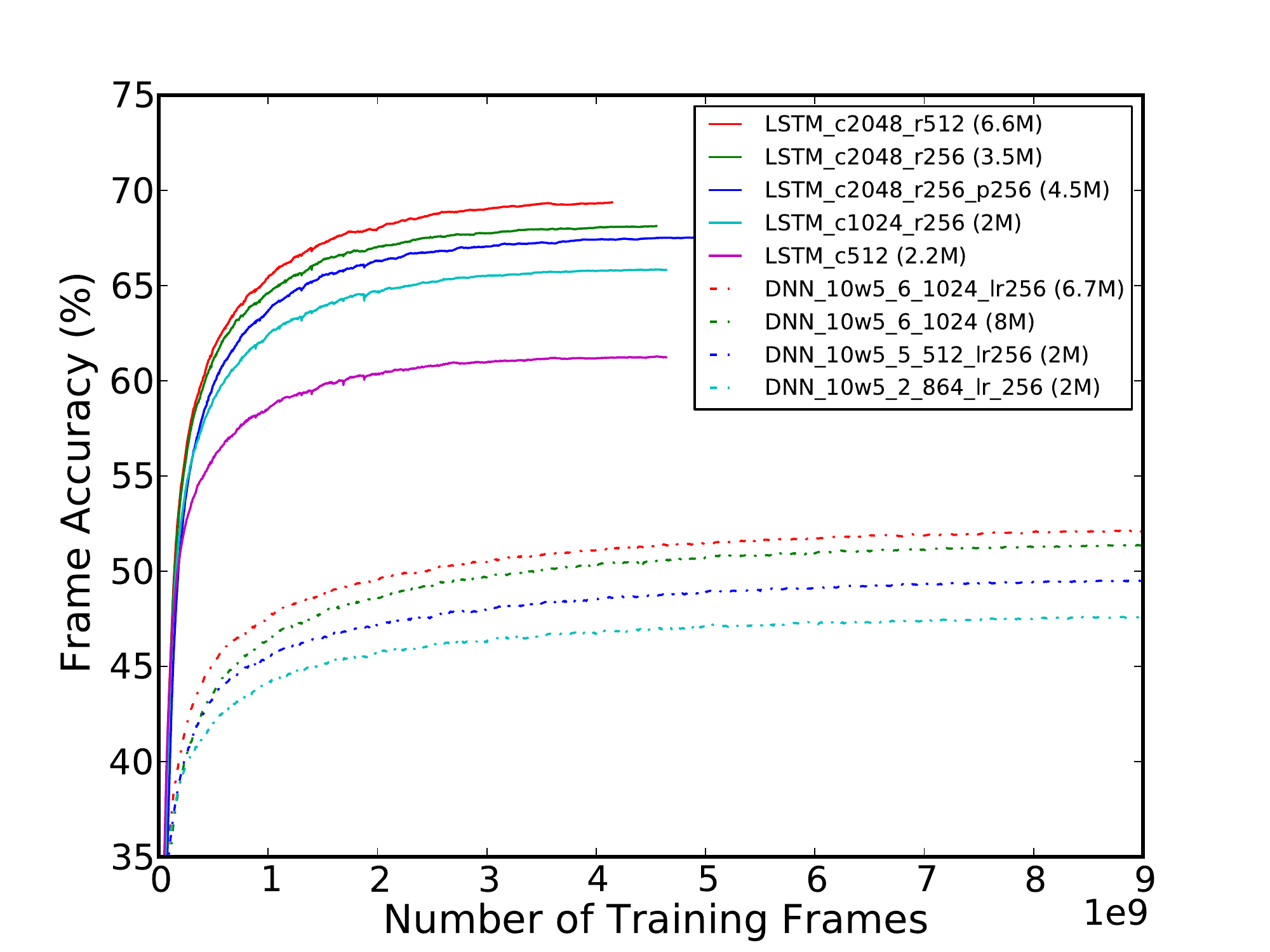}}
\caption{2000 context dependent phone HMM states.}
\label{fig:frame_acc2}
\end{figure}

\begin{figure}[ht]
  \centering
  \centerline{\includegraphics[width=8.5cm]{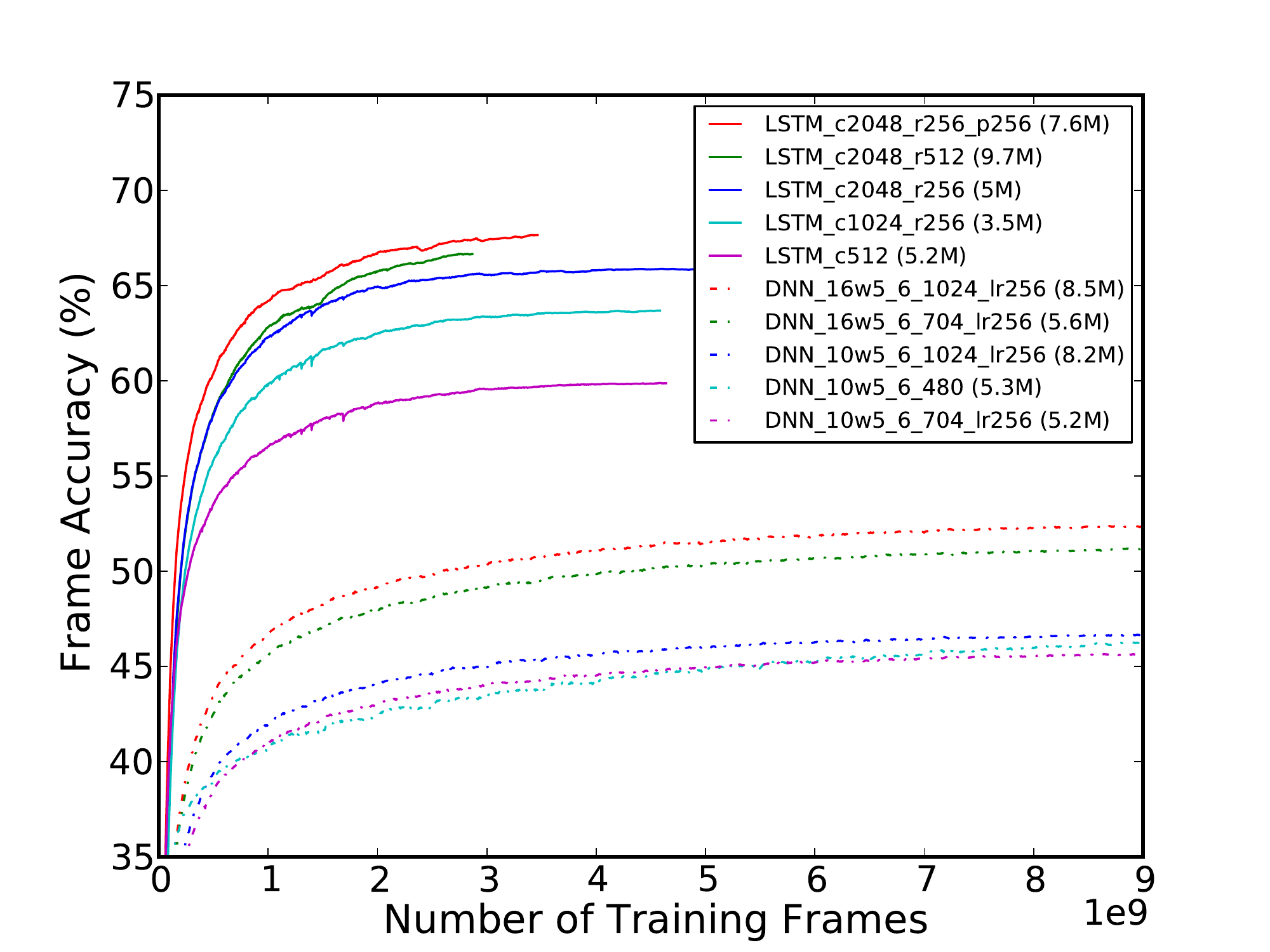}}
\caption{8000 context dependent phone HMM states.}
\label{fig:frame_acc3}
\end{figure}

Figure~\ref{fig:frame_acc1}, ~\ref{fig:frame_acc2}, and ~\ref{fig:frame_acc3} show the frame accuracy results for 126, 2000 and 8000 state outputs, respectively.
In the figures, the name of the network configuration contains the information about the network size and architecture.
$cN$ states the number ($N$) of memory cells in the LSTMs and the number of units in the hidden layer in the RNNs.
$rN$ states the number of recurrent projection units in the LSTMs and RNNs.
$pN$ states the number of non-recurrent projection units in the LSTMs.
The DNN configuration names state the left context and right context size (e.g. 10w5), the number of hidden layers (e.g. 6), the number of units in each of the hidden layers (e.g. 1024) and optional low-rank projection layer size (e.g. 256).
The number of parameters in each model is given in parenthesis.
We evaluated the RNNs only for 126 state output configuration, since they performed significantly worse than the DNNs and LSTMs.
As can be seen from Figure~\ref{fig:frame_acc1}, the RNNs were also very unstable at the beginning of the training and, to achieve convergence, we had to limit the activations and the gradients due to the exploding gradient problem.
The LSTM networks give much better frame accuracy than the RNNs and DNNs while converging faster.
The proposed LSTM projected RNN architectures give significantly better accuracy than the standard LSTM RNN architecture with the same number of parameters -- compare \textit{LSTM\_512} with \textit{LSTM\_1024\_256} in Figure~\ref{fig:frame_acc2}.
The LSTM network with both recurrent and non-recurrent projection layers generally performs better than the LSTM network with only recurrent projection layer except for the 2000 state experiment where we have set the learning rate too small.

\begin{figure}[ht]
  \centering
  \centerline{\includegraphics[width=8.5cm]{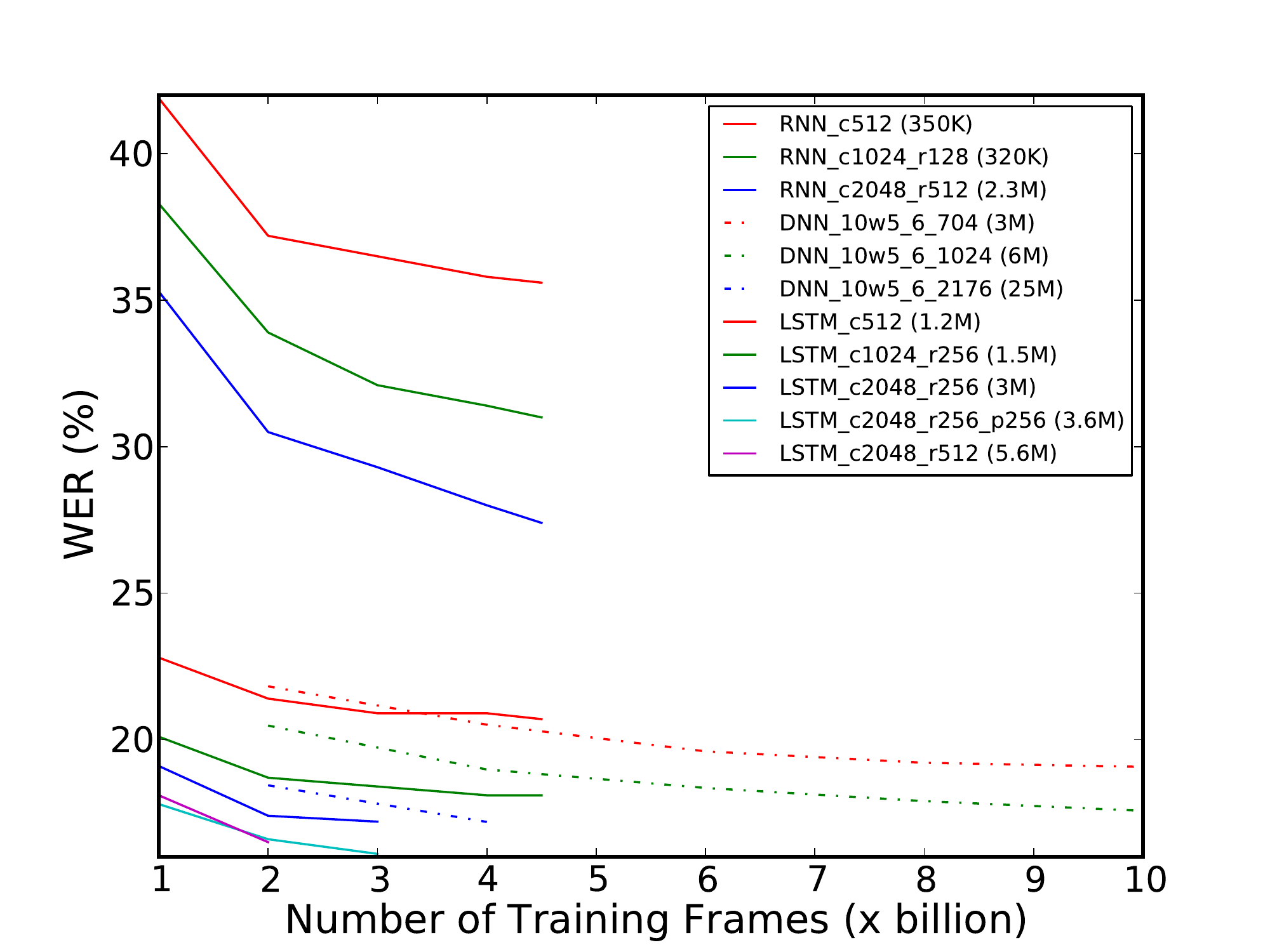}}
\caption{126 context independent phone HMM states.}
\label{fig:wer1}
\end{figure}

\begin{figure}[ht]
  \centering
  \centerline{\includegraphics[width=8.5cm]{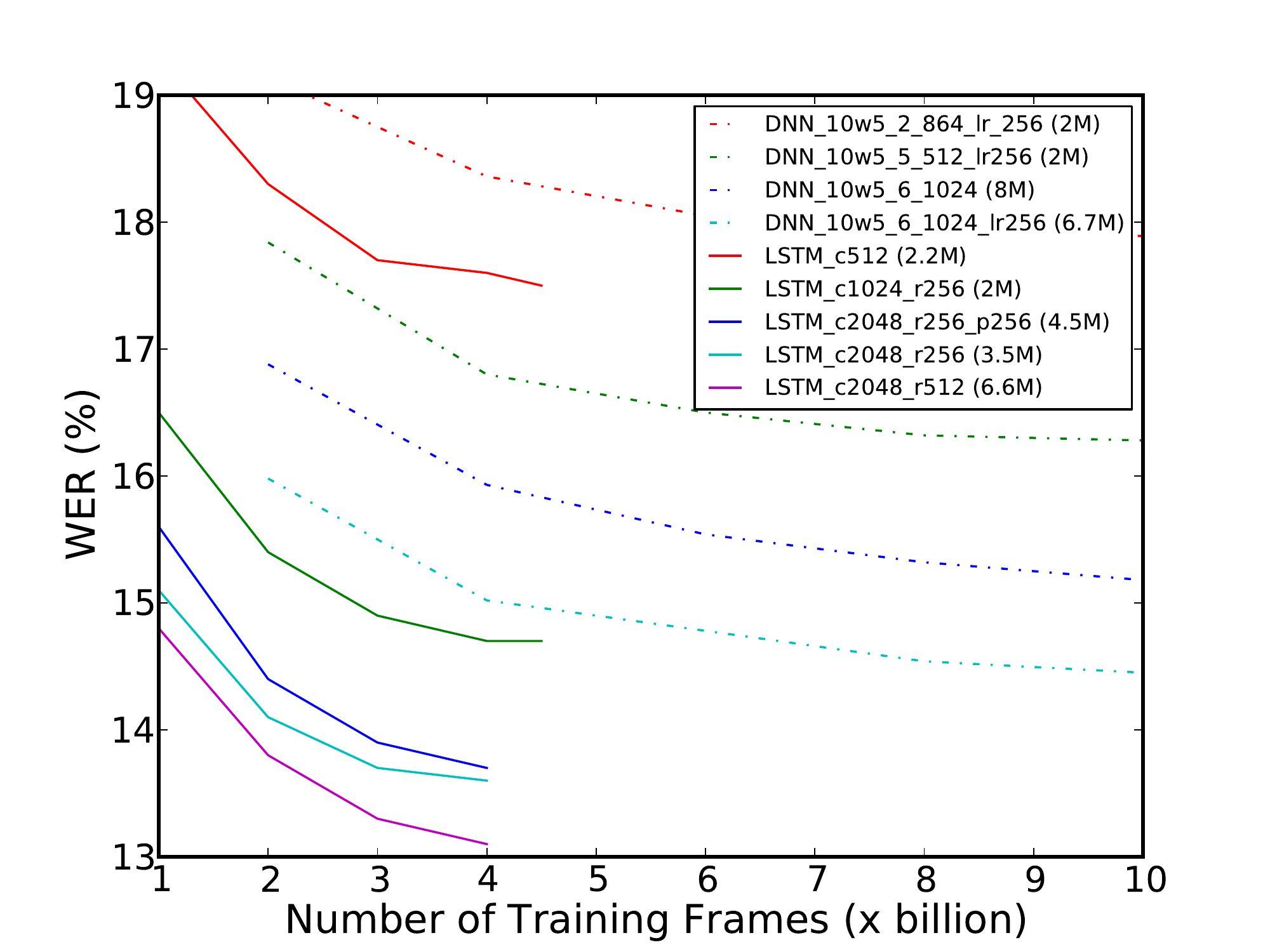}}
\caption{2000 context dependent phone HMM states.}
\label{fig:wer2}
\end{figure}

\begin{figure}[ht]
  \centering
  \centerline{\includegraphics[width=8.5cm]{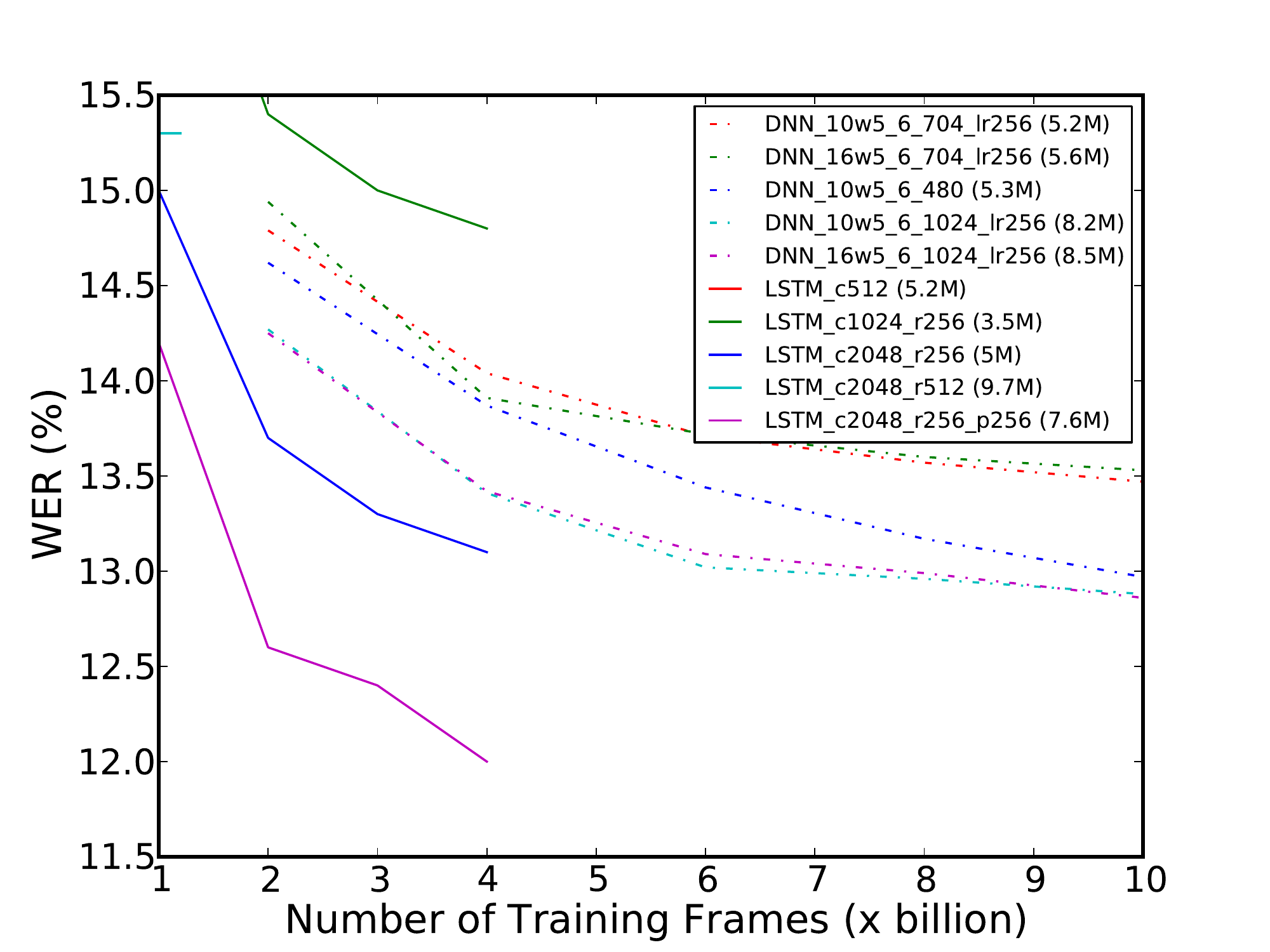}}
\caption{8000 context dependent phone HMM states.}
\label{fig:wer3}
\end{figure}

Figure~\ref{fig:wer1}, ~\ref{fig:wer2}, and ~\ref{fig:wer3} show the WERs for the same models for 126, 2000 and 8000 state outputs, respectively.
Note that some of the LSTM networks have not converged yet, we will update the results when the models converge in the final revision of the paper.
The speech recognition experiments show that the LSTM networks give improved speech recognition accuracy for the context independent 126 output state model, context dependent 2000 output state embedded size model (constrained to run on a mobile phone processor) and relatively large 8000 output state model.
As can be seen from Figure~\ref{fig:wer2}, the proposed architectures (compare \textit{LSTM\_c1024\_r256} with \textit{LSTM\_c512}) are essential for obtaining better recognition accuracies than DNNs.
We also did an experiment to show that depth is very important for DNNs -- compare \textit{DNN\_10w5\_2\_864\_lr256} with \textit{DNN\_10w5\_5\_512\_lr256} in Figure~\ref{fig:wer2}.

\section{Conclusion}
As far as we know, this paper presents the first application of LSTM networks in a large vocabulary speech recognition task.
To address the scalability issue of the LSTMs to large networks with large number of output units, we introduce two architecutures that make more effective use of model parameters than the standard LSTM architecture.
One of the proposed architectures introduces a recurrent projection layer between the LSTM layer (which itself has no recursion) and the output layer.
The other introduces another non-recurrent projection layer to increase the projection layer size without adding more recurrent connections and this decoupling provides more flexibility.
We show that the proposed architectures improve the performance of the LSTM networks significantly over the standard LSTM.
We also show that the proposed LSTM architectures give better performance than DNNs on a large vocabulary speech recognition task with a large number of output states.
Training LSTM networks on a single multi-core machine does not scale well to larger networks.
We will investigate GPU- and distributed CPU-implementations similar to~\cite{Dean:12} to address that.

\bibliographystyle{IEEEbib}
\bibliography{paper}

\begin{thebibliography}{10}

\bibitem{Bengio:94}
Yoshua Bengio, Patrice Simard, and Paolo Frasconi,
\newblock ``Learning long-term dependencies with gradient descent is
  difficult,''
\newblock {\em Neural Networks, IEEE Transactions on}, vol. 5, no. 2, pp.
  157--166, 1994.

\bibitem{Hochreiter:97}
Sepp Hochreiter and J\"{u}rgen Schmidhuber,
\newblock ``Long short-term memory,''
\newblock {\em Neural Computation}, vol. 9, no. 8, pp. 1735--1780, Nov. 1997.

\bibitem{Gers:00}
Felix~A. Gers, Jürgen Schmidhuber, and Fred Cummins,
\newblock ``Learning to forget: Continual prediction with {LSTM},''
\newblock {\em Neural Computation}, vol. 12, no. 10, pp. 2451--2471, 2000.

\bibitem{Gers:03}
Felix~A. Gers, Nicol~N. Schraudolph, and J\"{u}rgen Schmidhuber,
\newblock ``Learning precise timing with {LSTM} recurrent networks,''
\newblock {\em Journal of Machine Learning Research}, vol. 3, pp. 115--143,
  Mar. 2003.

\bibitem{Gers:01}
Felix~A. Gers and Jürgen Schmidhuber,
\newblock ``{LSTM} recurrent networks learn simple context free and context
  sensitive languages,''
\newblock {\em IEEE Transactions on Neural Networks}, vol. 12, no. 6, pp.
  1333--1340, 2001.

\bibitem{Schuster:97}
Mike Schuster and Kuldip~K. Paliwal,
\newblock ``Bidirectional recurrent neural networks,''
\newblock {\em Signal Processing, IEEE Transactions on}, vol. 45, no. 11, pp.
  2673--2681, 1997.

\bibitem{Graves:05}
Alex Graves and Jürgen Schmidhuber,
\newblock ``Framewise phoneme classification with bidirectional {LSTM} and
  other neural network architectures,''
\newblock {\em Neural Networks}, vol. 12, pp. 5--6, 2005.

\bibitem{Graves:09}
Alex Graves, Marcus Liwicki, Santiago Fernandez, Roman Bertolami, Horst Bunke,
  and J\"{u}rgen Schmidhuber,
\newblock ``A novel connectionist system for unconstrained handwriting
  recognition,''
\newblock {\em Pattern Analysis and Machine Intelligence, IEEE Transactions
  on}, vol. 31, no. 5, pp. 855--868, 2009.

\bibitem{Mohamed:12}
Abdel Rahman~Mohamed, George~E. Dahl, and Geoffrey~E. Hinton,
\newblock ``Acoustic modeling using deep belief networks,''
\newblock {\em IEEE Transactions on Audio, Speech \& Language Processing}, vol.
  20, no. 1, pp. 14--22, 2012.

\bibitem{Dahl:12}
George~E. Dahl, Dong Yu, Li~Deng, and Alex Acero,
\newblock ``Context-dependent pre-trained deep neural networks for
  large-vocabulary speech recognition,''
\newblock {\em IEEE Transactions on Audio, Speech \& Language Processing}, vol.
  20, no. 1, pp. 30--42, Jan. 2012.

\bibitem{Jaitly:12}
Navdeep Jaitly, Patrick Nguyen, Andrew Senior, and Vincent Vanhoucke,
\newblock ``Application of pretrained deep neural networks to large vocabulary
  speech recognition,''
\newblock in {\em Proceedings of INTERSPEECH}, 2012.

\bibitem{Graves:13}
Alex Graves, Abdel{-}rahman Mohamed, and Geoffrey Hinton,
\newblock ``Speech recognition with deep recurrent neural networks,''
\newblock in {\em Proceedings of ICASSP}, 2013.

\bibitem{Mikolov:10}
Tomáš Mikolov, Martin Karafiát, Lukáš Burget, Jan Černocký, and Sanjeev
  Khudanpur,
\newblock ``Recurrent neural network based language model,''
\newblock in {\em Proceedings of INTERSPEECH}. 2010, vol. 2010, pp. 1045--1048,
  International Speech Communication Association.

\bibitem{Dean:12}
Jeffrey Dean, Greg Corrado, Rajat Monga, Kai Chen, Matthieu Devin, Quoc~V. Le,
  Mark~Z. Mao, Marc'Aurelio Ranzato, Andrew~W. Senior, Paul~A. Tucker, Ke~Yang,
  and Andrew~Y. Ng,
\newblock ``Large scale distributed deep networks.,''
\newblock in {\em NIPS}, 2012, pp. 1232--1240.

\bibitem{Eigen:10}
Ga\"{e}l Guennebaud, Beno\^{i}t Jacob, et~al.,
\newblock ``Eigen v3,'' http://eigen.tuxfamily.org, 2010.

\bibitem{Williams:90}
Ronald~J. Williams and Jing Peng,
\newblock ``An efficient gradient-based algorithm for online training of
  recurrent network trajectories,''
\newblock {\em Neural Computation}, vol. 2, pp. 490--501, 1990.

\bibitem{Sainath:13}
T.N. Sainath, B.~Kingsbury, V.~Sindhwani, E.~Arisoy, and B.~Ramabhadran,
\newblock ``Low-rank matrix factorization for deep neural network training with
  high-dimensional output targets,''
\newblock in {\em Proc. ICASSP}, 2013.

\end{thebibliography}

\end{document}